\documentclass[a4paper, 10pt, conference]{ieeeconf}
\usepackage{FG2026}

\FGfinalcopy 

\IEEEoverridecommandlockouts
\usepackage{cite}
\usepackage{amsmath,amssymb,amsfonts}
\usepackage{algorithmic}
\usepackage{graphicx}
\usepackage{textcomp}

\usepackage{multirow}
\usepackage{array}
\usepackage{booktabs}
\usepackage{diagbox}
\usepackage{gensymb}
\usepackage{xcolor}
\usepackage{svg}

\def\FGPaperID{****} 

\title{\LARGE \bf
MoME: Estimating Psychological Traits from Gait with Multi-Stage \underline{M}ixture \underline{o}f \underline{M}ovement \underline{E}xperts
}

\author{\parbox{16cm}{\centering
    {\large Andy Cǎtrunǎ, Adrian Cosma, Emilian Rǎdoi}\\
    {\normalsize University Politehnica of Bucharest\\
    Faculty of Automatic Control and Computer Science\\
    {\tt\small \{andy\_eduard.catruna, ioan\_adrian.cosma, emilian.radoi\}@upb.ro}}}
}

\begin{document}

\ifFGfinal
\thispagestyle{empty}
\pagestyle{empty}
\else
\author{Anonymous FG2026 submission\\ Paper ID \FGPaperID \\}
\pagestyle{plain}
\fi
\maketitle

\begin{abstract}
Gait encodes rich biometric and behavioural information, yet leveraging the manner of walking to infer psychological traits remains a challenging and underexplored problem. We introduce a hierarchical Multi-Stage Mixture of Movement Experts (MoME) architecture for multi-task prediction of psychological attributes from gait sequences represented as 2D poses. MoME processes the walking cycle in four stages of movement complexity, employing lightweight expert models to extract spatio-temporal features and task-specific gating modules to adaptively weight experts across traits and stages. Evaluated on the PsyMo benchmark covering 17 psychological traits, our method outperforms state-of-the-art gait analysis models, achieving a 37.47\% weighted F1 score at the run level and 44.6\% at the subject level. Our experiments show that integrating auxiliary tasks such as identity recognition, gender prediction, and BMI estimation further improves psychological trait estimation. Our findings demonstrate the viability of multi-task gait-based learning for psychological trait estimation and provide a foundation for future research on movement-informed psychological inference.
\end{abstract}

\section{Introduction}
The pattern of human movement during walking has long been studied as a biometric identifier due to its uniqueness and variability across individuals \cite{pataky2012gait, park2021uniqueness}. Due to these factors, gait has become a topic of research in the area of deep learning and biometrics \cite{chao2019gaitset, fan2020gaitpart, lin2022gaitgl}, making gait analysis a mature topic that has been utilized in applications ranging from clinical diagnosis \cite{hulleck2022present,cosma2024psymo} to identification \cite{lin2022gaitgl,chao2019gaitset,catruna2024gaitpt,cosma2025scaling} and surveillance \cite{parashar2023real}, demonstrating the fact that human walking patterns encode a non-trivial amount of information about the subject. Unlike facial or vocal cues, gait can be captured unobtrusively at a distance and across diverse environments, making it particularly valuable for real-world applications where traditional modalities are not available. 

Psychological traits, encompassing personality attributes \cite{john1999big}, emotional states, and mental health indicators \cite{lovibond1995,ghq12}, have also become an important area of exploration within deep learning \cite{zhao2022deep, johannssen2018between}. Deep neural networks have been successfully utilized to predict psychological characteristics from facial images \cite{jeon2021deep, kachur2020assessing}, vocal samples \cite{rangra2023emotional}, videos \cite{gimeno2024videodepression}, or text data \cite{xue2018deep,10.1007/978-3-031-28244-7_13}. However, utilizing gait for the estimation of such traits has remained significantly underexplored, and previous attempts have had limited success due to the complexity and difficulty of the task \cite{cosma2024psymo}.

An increasing amount of research \cite{satchell2017evidence, fang2019depression,sun2017self} suggests that gait not only encodes physical and biomechanical properties but may also be an indicator of emotional or psychological states. This has been formulated in so-called theories of embodiment \cite{michalak2009embodiment,zatti2015embodied} which describe the two-way relationship between movement and the psyche: behaviour influences the mind and the mind influences behaviour and physiological responses. A simple example of this relationship is the fact that post-traumatic episodes results in elevated heart rate \cite{mcfarlane1994physical}, among others, and, conversely, hunger affects psychological states such as sadness and anxiety\cite{de2022associations}. As such, low-level movement features such as stride length, walking speed, and posture have been linked to traits like extraversion or aggression \cite{satchell2017evidence}. This hints that gait data may have the potential to be utilized for modeling, to a certain extent, internal psychological states \cite{cosma2024psymo}.

Utilizing gait to predict the psychological attributes of people entails multiple related tasks from partially overlapping movement cues. Multi-task learning is inherently difficult in this setting because of limited data: shared representations can entangle features, and prior work in multi-task learning \cite{zhao2018modulation} has observed destructive interference between tasks, where gradients for one objective suppress features needed by another, as tasks compete for capacity. Because of this, effective models for psychological attributes prediction from gait should: \textit{(i)} isolate task-relevant motion patterns when necessary, \textit{(ii)} enable selective sharing of features when helpful, and \textit{(iii)} adaptively allocate feature specialization across tasks and levels of movement complexity.

In this work, we propose a novel hierarchical \textbf{M}ixture \textbf{o}f \textbf{M}ovement \textbf{E}xperts (MoME) architecture, specifically designed for analyzing gait sequences for multi-task learning in the context of psychological traits estimation. Our architecture is designed hierarchically, consisting of multiple stages, with each stage dedicated to analyzing distinct levels of movement complexity. Within each stage, several movement experts process the information and become specialized in extracting different relevant motion features. Task-specific gating networks compute expert weights tailored for each psychological trait prediction task. The final prediction is aggregated from the task-specific gates across all stages, enabling the model to learn relevant movement data that is useful in most tasks as well as focus on task-specific features. 


The results on the PsyMo benchmark \cite{cosma2024psymo} for predicting 17 different psychological characteristics from gait data show that our model outperforms existing models that are popular in gait analysis. Furthermore, we investigate the impact of multi-task learning within this context and analyze how our architecture behaves when adding extra tasks. We find that utilizing additional tasks, such as re-identification, gender classification, and Body Mass Index (BMI) estimation contributes in extracting relevant movement signals and improves performance in the prediction of psychological traits. By analyzing the expert activations across stages and tasks we discover meaningful shared representations between psychological traits that align with current psychology findings \cite{ghq-fatigue, robins2001personality, stephan2022personality}.

The main contributions of our work are as follows:
\begin{enumerate}
    \item We propose a novel hierarchical multi-stage Mixture of Movement Experts (MoME) architecture tailored specifically for analyzing gait sequences at different levels of movement for multi-task learning and psychological traits estimation. Our architecture enables interpretability through expert activation analysis, revealing which tasks share representations and which require specialized or a combination of experts.
    \item We outperform current methods in psychological traits estimation from gait data. Our model obtains 37.47\% overall run-level F1 score on PsyMo and 44.6\% overall subject-level F1 score for predicting all attributes.
    \item We conduct an analysis on the efficacy of multi-task learning in the context of gait analysis, demonstrating that auxiliary tasks such as re-identification, gender classification and BMI estimation improves performance in predicting psychological attributes.
\end{enumerate}

\section{Related Work}
\subsection{Gait Analysis}
Gait analysis methods fall into two main categories: appearance-based \cite{chao2019gaitset, fan2020gaitpart, lin2022gaitgl, fan2023exploring} and model-based \cite{li2020jointsgait, cosma2022learning, catruna2024gaitpt, fan2024skeletongait}.

\textbf{Appearance-based} methods typically rely on sequences of human silhouettes extracted using background subtraction \cite{piccardi2004background} or segmentation techniques \cite{he2017mask, cheng2021per,kirillov2023segment}. CNNs are the most popular neural network in this category, with methods such as GaitSet \cite{chao2019gaitset,chao2021gaitset}, which represents gait sequences as unordered silhouette sets to force the model to learn the natural sequence of movements. Fan et al. \cite{fan2020gaitpart} proposed specialized convolutions focusing on distinct silhouette regions with different movement variations, whereas GLN \cite{hou2020gait} employs convolutional pyramids for compact representation. To simultaneously capture global and local gait features, Lin et al. introduced GaitGL \cite{lin2022gaitgl}, a two-stream network that addresses the limitations of previous methods which overlook hierarchical relations between features. Wang et al. \cite{wang2023dygait} construct a model that focuses on the dynamic parts of the body during movement which contain the relevant gait information and disregards less informative regions. Fan et al. \cite{fan2023exploring} propose a general method for building deep gait recognition networks, constructing basic blocks for both CNN-based and transformer-based architectures which are capable of extracting spatial and temporal features from sequences of silhouettes.

\textbf{Model-based} approaches focus mostly on human skeleton data obtained from pose estimation models \cite{xu2022vitpose, xu2023vitpose++} and from SMPL \cite{loper2023smpl} data which additionally encodes the shape of the body. Graph-based networks designed for action recognition, such as ST-GCN \cite{yan2018spatial} or ResGCN \cite{song2020stronger} have been successfully utilized for gait analysis. Li et al. proposed JointsGait \cite{li2020jointsgait} which applies ST-GCN to model spatio-temporal patterns, enhancing the representation power through joint relationship pyramid mapping. Teepe et al. \cite{teepe2021gaitgraph, teepe2022towards} similarly utilize graph convolutions with contrastive learning and data augmentation. Cosma and Radoi \cite{cosma2021wildgait} utilize surveillance data to pretrain ST-GCN in a self-supervised manner, making it robust to scenarios in the wild. Recent transformer-based architectures have started to be utilized in gait analysis. GaitFormer \cite{cosma2022learning} uses temporal attention on flattened skeletons, capturing the movement information through self-attention. GaitPT \cite{catruna2024gaitpt} improves on this approach, utilizing an anatomically informed pyramid vision transformer that extracts gait information at multiple levels of movement. SkeletonGait \cite{fan2024skeletongait} is constructed to work on skeleton maps which are silhouette-like images obtained from pose estimation data, whereas SkeletonGait++ \cite{fan2025opengait} utilizes both silhouettes and skeleton maps as input.

\subsection{Psychological Traits from Biometric Data}
Deep learning has increasingly been applied to infer psychological traits from biometric information, more specifically facial and gait data \cite{moreno2020estimation, xu2021prediction, kachur2020assessing, cosma2024psymo}. 

The study of psychological traits from biometric data with the help of deep learning is relatively underexplored, mostly because of the lack of data and because of the complexity of the task. Methods employing facial images and videos have shown that there are correlations between biometric data and the Big Five \cite{moreno2020estimation, kachur2020assessing, xu2021prediction}, the level of stress \cite{jeon2021deep, zhang2020video}. Furthermore, gait information has also been shown to be correlated with certain personality attributes or psychological states \cite{satchell2017evidence, sun2017self, fang2019depression}.

The Big Five personality traits have been among the most utilized for prediction psychological attributes from facial features. Moreno et al. \cite{moreno2020estimation} treat the task of psychological traits prediction as a binary classification task on the BFI, utilizng a single face image as input to the CNN model. In contrast, Katchur et al. \cite{kachur2020assessing} approach it as a regression task, obtaining the highest performance on conscientiousness. Meanwhile, Xu et al. \cite{xu2021prediction} found that their multi-class classification strategy with a CNN model obtains the highest accuracy on neuroticism and extraversion.

Facial images have also been utilized to detect the stress level. For example, Jeon et al. \cite{jeon2021deep} employ an architecture that combines a CNN with temporal attention mechanism to weight the features of frames with highest stress levels. Their approach utilizes as input RGB video data as well as facial landmarks for each frame. On the other hand, Zhang et al. \cite{zhang2020video} utilize a hybrid model, employing CNNs with LSTMs and attention in order to detect the stress level, treating it as a classification task.

Beyond full-face analysis, ocular region data consisting of images of the eye region has also been explored for analyzing the psychological state. This ocular area has been shown to be linked with the cognitive load of the person. Rahman et al. \cite{rahman2021vision} experiment with both CNN and SVM models on eye movement data, treating the problem as a classification. Abilkassov et al. \cite{abilkassov2021system} design a system that employs a CNN model on face images in order to predict the cognitive load.

Several gait features have been linked with certain psychological attributes. For instance, Satchell et al. \cite{satchell2017evidence} found that the walking speed of the person along with the magnitude of the upper body movement and lower body movement are associated with certain Big Five personality traits as well as aggression levels. Fang et al. \cite{fang2019depression} have shown that depressed people exhibit certain gait abnormalities in terms of walking speed, stride length, arm swing and other movement features when compared to non-depressed people. Sun et al. \cite{sun2017self} employed standard machine learning models to reveal strong correlations between particular gait features and self-esteem.

Building upon these findings, The PsyMo dataset \cite{cosma2024psymo} has been proposed as a method for measuring the capabilities of deep learning models to capture the relationship between psychological traits and gait information. The dataset includes walking sequences from 312 participants recorded under distinct walking conditions and from different camera perspectives. Alongside gait recordings, participants completed standardized psychological questionnaires, obtaining 17 psychometric attributes such as personality, self-esteem, fatigues, aggressiveness, and mental health. The dataset provides 2D and 3D pose estimation skeletons, 3D SMPL meshes and silhouettes. We utilize the 2D skeletons of this dataset both for training and evaluating our proposed architecture.


\section{Method}

\begin{figure}[hbt!]
\centering
\includesvg[width=0.9\linewidth]{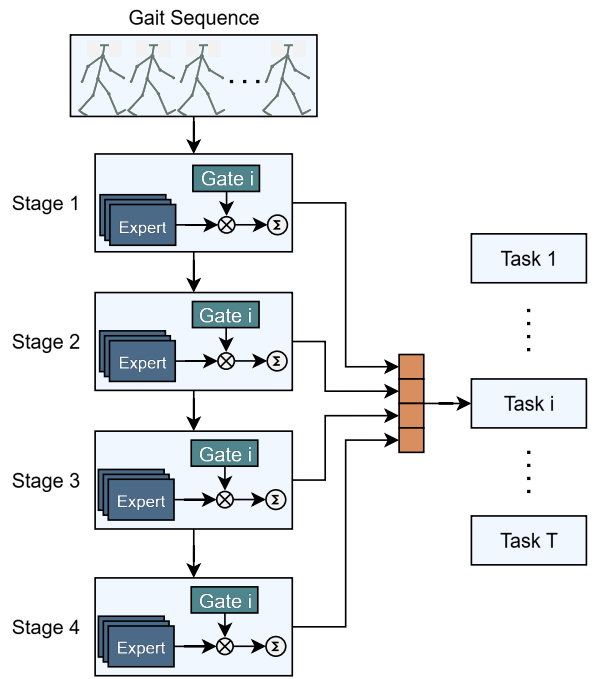}
\caption{High-level overview of our proposed method consisting of a multi-stage Mixture of Movement Experts framework. At each stage, representing a different level of movement complexity, every task has an assigned gating module which computes the optimal weight of the expert outputs for making the prediction.}
\label{fig:overview}
\end{figure}

In this section, we present our approach to predicting multiple personality attributes from walking patterns. We frame this task as a multi-task classification in which there are a number of different psychological attributes to predict. Our proposed method, shown in Figure \ref{fig:overview}, utilizes multi-stage multi-gate experts that specialize in extracting task-specific relevant movement patterns. Our model builds upon the family of Mixture of Experts (MoE) architectures \cite{eigen2013learning, shazeer2017outrageously}, which employ independent expert modules whose outputs are adaptively weighted by learned gating mechanisms, enabling specialization.

\subsection{Multi-Task Problem Setup}
Given a person's walking pattern represented as a sequence of pose estimation skeletons, the goal is to predict a set of psychological attributes of the walking person. Each psychological attribute is associated with multiple labels and is treated as a classification task. Consequently, the problem of predicting various psychological labels from the same gait input can be regarded as a multi-task learning problem.

Formally, let $X \in \mathbb{R}^{n \times j \times c}$ denote the sequence of pose estimation skeletons, where $n$ is the temporal dimension (number of poses/frames) , $j$ is the number of structural units (joints/parts), and $c$ is the channel dimensionality (i.e., $c=2$ for 2D poses). The goal is to learn a neural network $f_W(X)$ that outputs a set of predicted labels $\{y_1, y_2,...,y_T\}$, where each $y_t$ is the predicted class for task $t \in \{1,...,T\}$. In particular, in our context, each task $t$ can have a different number of classes $C_t$, ranging from 3 to 5.

In this setup, the problem is that different tasks may require different types of features at various granularity levels. In particular, each psychological attribute may require a distinct combination of features, depending on the nature and complexity of the trait. This can lead to competition for shared features during training, where tasks with differing feature requirements interfere with one another. If the model does not effectively route information correctly for each task, it could result in some tasks or movement features dominating the learning process. 

\subsection{Multi-Stage Mixture of Movement Experts}
\begin{figure}[hbt!]
    \centering
    \includesvg[width=\linewidth]{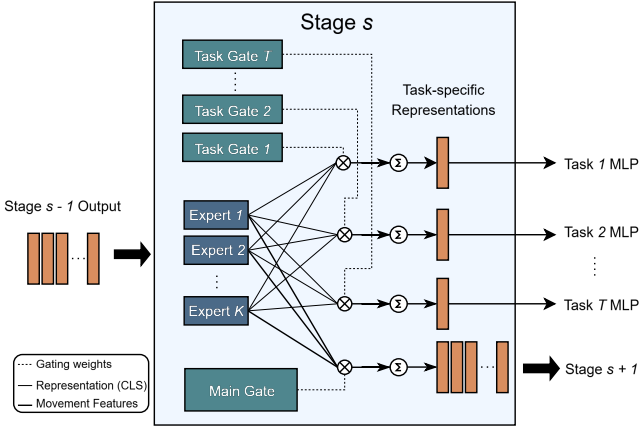}
    \caption{Visualization of single stage module. Movement features obtained from a previous stage are processed by multiple experts to obtain more complex features. Independent Task Gates weight each of the expert outputs for obtaining task-specific representations while the \textit{Main Gate} obtains the weights of the movement features for the next stage.}
    \label{fig:stage-module}
\end{figure}

To address this problem, we propose a multi-stage \textbf{M}ixture \textbf{o}f \textbf{M}ovement \textbf{E}xperts (MoME) architecture, as shown in Figure \ref{fig:overview}. Our model leverages the inherent stage-wise structure of GaitPT \cite{catruna2024gaitpt}, where each stage focuses on different levels of movement representation: Stage 1 deals with joint-level features, Stage 2 focuses on limb-level features, Stage 3 extracts limb-group features, while Stage 4 computes body-level features. MoME expands on this approach, incorporating Stage experts and per-task gate modules, allowing the network to be more expressive.

We propose the following architectural components: 1) \textit{Stage Experts} that learn independently to extract different stage-specific relevant features, 2) \textit{Main Gate} which decides what movement information to retain and discard from stage to stage and 3) \textit{Task-Specific Gates} at each stage to allow each task to decide how much emphasis to place on each expert's output and level of movement. The interaction between these components is visualized in Figure \ref{fig:stage-module}.

\begin{figure}[hbt!]
    \centering
    \includesvg[width=\linewidth]{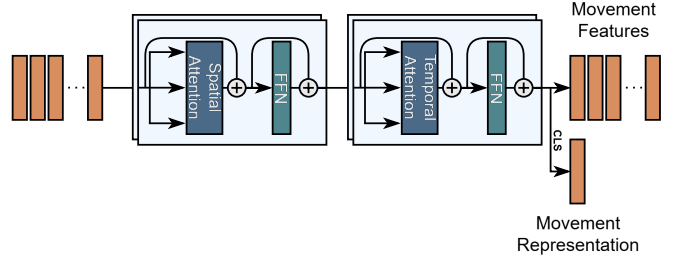}
    \caption{Visualization of the expert and gating module architectures. Following the hierarchical design of GaitPT \cite{catruna2024gaitpt} our expert and gating modules are designed to extract relevant spatio-temporal features by utilizing an encoder with spatial attention and one with temporal attention.}
    \label{fig:expert}
\end{figure}

The three proposed components have the same internal structure, shown in Figure \ref{fig:expert}, employing self-attention and MLPs at different levels of movement in a similar fashion to the building block of GaitPT. \textit{Experts} and \textit{Gates} can differ in size and number of parameters based on an amplifier hyperparameter which multiplies the depth, embedding size and number of self-attention heads. 

\textbf{Stage Experts.} At each stage $s \in \{1,2,3,4\}$, we define a set of $K_s$ expert modules $\{E_{s,1}, E_{s,2}, \ldots, E_{s,K_s} \}$ where $K_s$ is a hyperparameter for the number of experts at stage $s$ and $E_{s,i}$ is a lightweight expert responsible for extracting relevant features at the current stage $s$. The module utilizes self-attention to extract spatio-temporal information based on the specific level of the stage, in a hierarchical manner.

Each \textit{Expert} has the same structure and number of parameters, but during training they specialize in extracting distinct movement features. During inference, these diverse representations are combined through task-specific gates to enhance performance.

Formally, let $\mathbf{H}_{s-1} \in \mathbb{R}^{n \times j_{s-1} \times c_{s-1}}$ denote the representation passed from the previous stage (with $\mathbf{H}_0 = X$ being the raw input). The number of structural parts $j_s$ decreases from stage to stage while the channel dimensionality $c_s$ increases (e.g., when merging features corresponding to joints into features corresponding to limbs). Each \textit{Expert} $E_{s,k}$ transforms $\mathbf{H}_{s-1}$ into some stage-specific feature representation:
\begin{equation}
    \mathbf{Z}_{s,k} = E_{s,k}(\mathbf{H}_{s-1})    
\end{equation}

We then use a gating mechanism (\textit{Main Gate}) to aggregate these outputs into the stage output $\mathbf{H}_s$ that will be the input for the next stage.

\textbf{Main Gate.} Information is propagated to the next stage with the help of a module we refer to as \textit{Main Gate}. This module assigns a weight to each experts in the current stage, and the output of the current stage is computed as a weighted sum of their results, which then serves as input to the next stage.

Formally, the \textit{Main Gate} $G_s$ processes the previous stage representation $\mathbf{H}_{s-1}$ to compute gating weights for the current stage experts. Given the set of expert outputs ${\mathbf{Z}_{s,1}, \mathbf{Z}_{s,2}, \dots, \mathbf{Z}_{s,K_s}}$, the \textit{Main Gate} computes a weighting vector as follows:

\begin{equation}
\mathbf{\alpha_s} = \text{softmax}\left(\mathbf{G_s}(\mathbf{H}_{s-1})\right),
\end{equation}

where $\mathbf{\alpha}_s \in \mathbb{R}^{K_s}$ represents the normalized importance of each expert’s output at stage $s$.

Finally, the aggregated output representation passed to the following stage is computed as a weighted combination of expert outputs:
\begin{equation}
\mathbf{H_s} = \sum_{k=1}^{K_s} \mathbf{\alpha}_{s,k} \mathbf{Z}_{s,k},
\end{equation}
where $\mathbf{\alpha}_{s,k}$ denotes the gating weight assigned to the $k{\text{th}}$ expert at stage $s$. This weighted aggregation ensures that the most relevant expert information is emphasized and retained through the full architecture.

\textbf{Task-Specific Gates.} To support multi-task learning, the architecture employs task-specific gates, each tailored to a particular task and processing information from a specific stage. Each task $t$ has a dedicated gating mechanism at every stage $s$, resulting in four task-specific gates per task, each with a different level of movement information. Similarly to the \textit{Main Gate}, a \textit{Task-Specific Gate} computes a set of weights based on the previous stage representation $\mathbf{H}_{s-1}$, determining the relevance of each expert output for its task.

Differently from the \textit{Main Gate}, the task-gate predictions are utilized to weight the summarized movement representation of the experts. This summarized representation is computed as the \texttt{CLS} vector $\mathbf{z}_{s,k}^{\text{\texttt{CLS}}} \in \mathbb{R}^{C_s}$ of the extracted movement features $\mathbf{Z}_{s,k}$.

Formally, given task $t$ and stage $s$, the \textit{Task-Specifc Gate} $G_s^t$ computes the weights as follows:
\begin{equation}
\mathbf{\alpha}_s^t = \text{softmax}\left(G_s^t(\mathbf{H}_{s-1})\right),
\end{equation}
where $\mathbf{\alpha}_s^t \in \mathbb{R}^{K_s}$ represents the normalized importance of each expert's output specifically for task $t$ at stage $s$.

The task-specific stage representation $\mathbf{h}_s^t \in \mathbb{R}^{C_s}$ is then obtained through a weighted sum of expert summarized representations:
\begin{equation}
\mathbf{h}_s^t = \sum_{k=1}^{K_s} \mathbf{\alpha}_{s,k}^t \mathbf{z}_{s,k}^{\text{\texttt{CLS}}},
\end{equation}
where $\mathbf{\alpha}_{s,k}^t$ denotes the weight assigned by the task-specific gate to the $k{\text{th}}$ expert at stage $s$ for task $t$. This multi-stage multi-expert approach enables the model to dynamically adapt and prioritize expert features based on both hierarchical level of movement and the specific task it is assigned to.

\textbf{Final Output.} For each task $t$, the final output is computed by aggregating task-specific representations from all stages. Specifically, task-specific representations from each stage are concatenated to form a unified representation:
\begin{equation}
\mathbf{h}^t = \text{concat}(\mathbf{h}_1^t, \mathbf{h}_2^t, \mathbf{h}_3^t, \mathbf{h}_4^t).
\end{equation}
This unified representation is then passed through an MLP comprising a hidden layer with a GELU \cite{gelu} nonlinearity, followed by an output layer. Formally, the output is computed as:
\begin{equation}
\mathbf{o}^t = \text{MLP}_t(\mathbf{h}^t),
\end{equation}
where $\mathbf{o}^t$ are logits in the context of classification or a single scalar in the case of regression tasks.

\begin{table*}[hbt]
\caption{Run-Level results for psychological traits estimation. We utilize weighted F1 scores as the evaluation metrics for all models. For our model, we show the mean and standard deviation across 3 training runs. Our proposed architecture achieves the highest overall performance. Table adapted from \cite{cosma2024psymo}.}
\label{tab:run-level}
\centering
\resizebox{\textwidth}{!}{%
\begin{tabular}{l|ccccc|c|cccc|ccc|ccc|c|c}
\toprule
& \multicolumn{5}{c}{\textbf{BFI (4-classes)}} 
& \multicolumn{1}{c}{\textbf{RSE (3-classes)}} 
& \multicolumn{4}{c}{\textbf{BPAQ (4-classes)}} 
& \multicolumn{3}{c}{\textbf{OFER (4-classes)}} 
& \multicolumn{3}{c}{\textbf{DASS (5-classes)}} 
& \multicolumn{1}{c|}{\textbf{GHQ (3-classes)}} 
& \textbf{Overall} \\
\cmidrule(r){2-6}
\cmidrule(r){7-7}
\cmidrule(r){8-11}
\cmidrule(r){12-14}
\cmidrule(r){15-17}
\cmidrule(r){18-18}
\textbf{Method} 
& O & C & E & A & N
& Esteem
& Phys. & Verbal & Anger & Host.
& Chronic & Acute & Recov.
& Depr. & Anxiety & Stress
& GHQ & \textbf{Mean} \\
\midrule
GaitFormer \cite{cosma2022learning}
& 23.80 & 25.35 & 27.88 & 28.12 & 28.21
& 43.59
& 28.27 & 31.89 & \textbf{31.24} & 28.84
& 27.43 & 33.05 & 34.83
& 28.27 & 24.55 & 43.69
& 49.38
& 31.67 \\

GaitGraph \cite{teepe2021gaitgraph}
& 22.10 & 24.89 & 26.39 & 26.09 & 23.67
& 45.49
& 27.86 & 24.21 & 24.24 & 27.83
& 25.17 & 29.61 & 30.00
& 30.82 & 19.45 & 42.73
& 52.94
& 29.62 \\

GaitSet \cite{chao2019gaitset}
& 26.60 & 26.51 & 25.81 & 26.87 & 27.89
& 45.40
& 26.51 & 29.93 & 27.71 & 28.81
& 31.13 & 33.14 & 34.80
& 42.20 & 27.57 & 51.98
& 56.11
& 33.46 \\

GaitGL \cite{lin2022gaitgl}
& 26.64 & 23.48 & 25.70 & 25.94 & 28.30
& 44.73
& 26.28 & 29.99 & 28.16 & 27.90
& 30.35 & 31.92 & 32.95
& 37.64 & 25.03 & 46.03
& 54.79
& 32.10 \\

SMPLGait \cite{zheng2022gait}
& 24.14 & 20.83 & 23.21 & 25.75 & 32.21
& 45.90
& 28.32 & 28.09 & 24.13 & 30.18
& 30.07 & 34.54 & 35.05
& 42.74 & 21.91 & 51.67
& 52.22
& 32.40 \\

GaitPT \cite{catruna2024gaitpt}
& 25.17 &  25.3 &  26.78 &  27.25 &  30.28 &  52.43 &  24.61 &  30.07 &  26.32 &  \textbf{35.04} &  34.65 &  37.41 &  34.99 &  42.45 &  27.71 &  53.4 &  56.97 &  34.75 \\

\midrule

\textbf{MoME (Ours)} 
&
\textbf{27.14} \scriptsize{$\pm$0.6} & \textbf{27.87} \scriptsize{$\pm$0.7} & \textbf{28.47} \scriptsize{$\pm$0.7} & \textbf{30.76} \scriptsize{$\pm$0.3} & \textbf{32.97} \scriptsize{$\pm$1.0} & \textbf{53.17} \scriptsize{$\pm$0.2} & \textbf{29.83} \scriptsize{$\pm$1.0} & \textbf{35.68} \scriptsize{$\pm$0.3} & 31.01 \scriptsize{$\pm$0.5} & 34.75 \scriptsize{$\pm$1.3} & \textbf{38.36} \scriptsize{$\pm$0.9} & \textbf{39.97} \scriptsize{$\pm$1.5} & \textbf{39.52} \scriptsize{$\pm$2.5} & \textbf{43.54} \scriptsize{$\pm$0.7} & \textbf{31.42} \scriptsize{$\pm$0.5} & \textbf{54.78} \scriptsize{$\pm$0.8} & \textbf{57.73} \scriptsize{$\pm$0.4} & \textbf{37.47} \scriptsize{$\pm$0.1} \\

\bottomrule
\end{tabular}%
}
\end{table*}

Our proposed approach allows for more expressivity in the model as it can isolate task-specific features from each level of movement while mixing and matching more general features through all paths in the architecture.

\subsection{Implementation Details}
\begin{table*}[hbt]
\caption{Subject-Level results for psychological traits estimation. We utilize weighted F1 scores as the evaluation metric. For our model, we show the mean and standard deviation across 3 training runs. Aggregating the prediction from multiple viewpoints and scenarios reliably improves performance on all tasks for our proposed model. Table adapted from \cite{cosma2024psymo}.}
\label{tab:subject-level}
\centering
\resizebox{\textwidth}{!}{%
\begin{tabular}{l|ccccc|c|cccc|ccc|ccc|c|c}
\toprule
& \multicolumn{5}{c}{\textbf{BFI (4-classes)}} 
& \multicolumn{1}{c}{\textbf{RSE (3-classes)}} 
& \multicolumn{4}{c}{\textbf{BPAQ (4-classes)}} 
& \multicolumn{3}{c}{\textbf{OFER (4-classes)}} 
& \multicolumn{3}{c}{\textbf{DASS (5-classes)}} 
& \multicolumn{1}{c|}{\textbf{GHQ (3-classes)}} 
& \textbf{Overall} \\
\cmidrule(r){2-6}
\cmidrule(r){7-7}
\cmidrule(r){8-11}
\cmidrule(r){12-14}
\cmidrule(r){15-17}
\cmidrule(r){18-18}
\textbf{Method} 
& O & C & E & A & N
& Esteem
& Phys. & Verbal & Anger & Host.
& Chronic & Acute & Recov.
& Depr. & Anxiety & Stress
& GHQ & \textbf{Mean} \\
\midrule
GaitFormer \cite{cosma2022learning}
& 33.12 & 31.03 & 36.27 & 38.42 & 33.25
& 55.58
& 38.00 & \textbf{42.81} & \textbf{42.73} & 39.09
& 36.71 & 38.88 & \textbf{46.80}
& 38.45 & 36.33 & 52.85
& 57.21
& 41.03 \\

GaitGraph \cite{teepe2021gaitgraph}
& 24.40 & 28.91 & 27.55 & 31.33 & 27.55
& 50.95
& 34.81 & 23.76 & 26.51 & 32.37
& 28.19 & 28.30 & 25.08
& 35.88 & 19.94 & 48.01
& 53.10
& 32.15 \\

GaitSet \cite{chao2019gaitset}
& 27.50 & 31.55 & 26.55 & 27.13 & 27.37
& 48.38
& 32.37 & 36.46 & 30.94 & 30.87
& 32.99 & 33.13 & 36.70
& 46.64 & 33.53 & 55.05
& 60.59
& 36.33 \\

GaitGL \cite{lin2022gaitgl}
& 31.04 & 23.78 & 28.05 & 23.15 & 30.73
& 48.35
& 26.89 & 41.79 & 30.37 & 28.80
& 34.22 & 36.96 & 37.31
& 46.20 & 27.75 & 55.57
& 60.54
& 35.97 \\

SMPLGait \cite{zheng2022gait}
& 22.77 & 18.02 & 19.74 & 26.80 & 36.51
& 48.62
& 33.80 & 33.42 & 20.47 & 34.89
& 31.83 & 37.27 & 38.71
& 42.66 & 21.89 & 52.72
& 58.97
& 34.06 \\

GaitPT \cite{catruna2024gaitpt}
& \textbf{38.52} &  25.01 &  \textbf{40.86} &  31.79 &  \textbf{39.16} &  49.87 &  26.22 &  41.36 &  32.07 &  \textbf{45.77} &  36.89 &  47 &  40.09 &  41.84 &  35.17 &  55.41 &  61.3 &  40.49 \\

\midrule

\textbf{MoME (Ours)}
& 34.18 \scriptsize{$\pm$1.6} & \textbf{36.65} \scriptsize{$\pm$2.6} & 34.66 \scriptsize{$\pm$0.6} & \textbf{40.33} \scriptsize{$\pm$3.1} & 38 \scriptsize{$\pm$0.6} & \textbf{59.95} \scriptsize{$\pm$0.6} & \textbf{39.32} \scriptsize{$\pm$3.1} & 41.61 \scriptsize{$\pm$1.2} & \textbf{42.85} \scriptsize{$\pm$1.0} & 42.01 \scriptsize{$\pm$2.5} & \textbf{47.67} \scriptsize{$\pm$2.1} & \textbf{47.39} \scriptsize{$\pm$0.3} & 42.82 \scriptsize{$\pm$1.6} & \textbf{47.45} \scriptsize{$\pm$0.6} & \textbf{42.37} \scriptsize{$\pm$3.7} & \textbf{57.59} \scriptsize{$\pm$0.0} & \textbf{63.33} \scriptsize{$\pm$1.6} & \textbf{44.6} \scriptsize{$\pm$0.5} \\

\bottomrule
\end{tabular}%
}
\end{table*}

We utilize the PsyMo dataset \cite{cosma2024psymo} for all of our experiments. The dataset consists of gait videos collected from 312 participants with 6 different camera angles along 7 walking variations. The walking variations include: normal walking (NM), walking while wearing a coat (CL), walking while carrying bag (BG), walking slower (WSS), walking faster (WSF), walking while texting (TXT), walking while talking on the phone (PH). We employ sequences of consecutive 2D pose estimation skeletons provided by the dataset as the gait input to our model.

The dataset provides psychological traits annotations for each participant obtained from various questionnaires. Each subject has scores for: Big Five Inventory (BFI) \cite{john1999big}, Rosenberg Self-Esteem (RSE) \cite{Rosenberg+2015}, Buss-Perry Aggression Questionnaire (BPAQ) \cite{buss1992aggression}, Occupational Fatigue Exhaustion/Recovery Scale (OFER) \cite{winwood2005development}, Depression, Anxiety and Stress Scale (DASS-21) \cite{lovibond1995}, General Health Questionnaire (GHQ) \cite{ghq12}. Most of these questionnaires contain multiple subscales. For example, DASS-21 measures Depression, Anxiety, and Stress. The scores can range from 1 to 3, 1 to 4, or 1 to 5 depending on the questionnaire. Additionally, the dataset contains other information about the participants, such as the gender and BMI.

We utilize the same train-test split as in the official partitioning in which the first 250 subjects are placed in the training set, while the remaining participants are utilized for evaluation. We treat each individual scale of the questionnaires as a classification task where the model has to predict the probability of each score. Therefore, we utilize the cross-entropy objective for training the model for psychological traits estimation as well as gender classification. We also train the model for identity recognition utilizing the triplet loss \cite{triplet-loss} and for BMI prediction with mean square error objective. In total, we train our model for 20 tasks utilizing a combined loss for all objectives and employing the AdamW optimizer \cite{kigma2014adam}.

To encourage expert specialization and prevent under-utilization, we incorporate 2 auxiliary losses. A load balancing loss penalizes deviations from uniform average expert usage across tasks, preventing scenarios where some experts are never utilized and others are utilized too much. An entropy regularization loss minimizes the entropy of each gate's output distribution, encouraging peaked expert selections per task. Together, these losses ensure all experts are utilized while enforcing task-specific specialization.

Our top-performing architecture utilizes 8 experts per stage, where the experts and \textit{Main Gate} module have the same number of parameters, while the task gates have half as many parameters. The full model is composed of 10M parameters and is trained with a cyclic learning rate schedule \cite{cyclical-lr} with a base learning rate value of 0.0001 and a maximum learning rate value of 0.0009, a step size of 25 epochs, and an exponential range policy with a decay factor of 0.999 for a total of 1000 training epochs. We utilize a batch size equal to 288 and a gait length of 55 consecutive pose estimation skeletons. The duration of a full training run is approximately 30 hours on an A100 GPU.

\section{Results}
We follow the official evaluation protocol of PsyMo for analyzing the performance of our proposed model for psychological traits estimation. We train the model on the samples with IDs ranging from 0 to 250 and evaluate on the remaining individuals. The PsyMo dataset proposes 2 types of evaluation for psychological traits: run-level and subject-level. Run-level evaluation implies making predictions for each separate walking sample, while subject-level evaluation implies making predictions based on all available samples of an individual. For the subject-level evaluation we utilize a simple strategy: we average over the predicted post-softmax class probabilities for all samples of an individual. We train both GaitPT \cite{catruna2024gaitpt} and our proposed model while utilizing the results reported in the PsyMo benchmark \cite{cosma2024psymo} for the other architectures.

\subsection{Psychological Attributes from Gait}
Tables \ref{tab:run-level} and \ref{tab:subject-level} show the performance of our model across both types of evaluation, comparing it to other widely utilized models for gait analysis. Similarly to our proposed model, GaitFormer and GaitGraph operate on sequences of 2D pose estimation skeletons. On the other hand, GaitSet and GaitGL employ silhouettes and SMPLGait utilizes 3D human meshes.

As noted in Table \ref{tab:run-level} our proposed model consistently outperforms all other gait analysis models in run-level psychological traits estimation, obtaining an overall weighted F1 score of 37.47, surpassing the closest model, GaitPT (34.75), by an absolute gain of 2.72. Against the strongest competing method in each subtask, our model obtains the most notable performance improvements on Occupational Fatigue Exhaustion Recovery (OFER), with +4.53 in Recovery, +3.71 in Chronic Fatigue, and +2.56 in Acute Fatigue.


We observe that all models underperform on the BFI and BPAQ subscales which points to the fact that these traits may be too difficult to predict from isolated gait samples. Within BFI, our model leads on all traits, but by small margins. For BPAQ, our proposed architecture obtains the highest performance on Physical Aggression (29.83), Verbal Aggression (35.68), but GaitFormer closely outperforms it on Anger (31.24 vs 31.01) and GaitPT narrowly surpasses it on Hostility (35.04 vs 34.75).

Our model builds upon GaitPT which utilizes a hierarchical approach to movement analysis, processing the motion in 4 stages with increasing complexity. The results show that this hierarchical approach consistently outperforms the other methods. The effect is most pronounced on classifying fatigue (OFER Chronic and Acute), on mental health traits (RSE Esteem, DASS Stress) where both GaitPT and MoME obtain a substantial gain over the other approaches to gait analysis (e.g., 53.17 and 52.43 vs 45.40 on RSE Esteem). 

Table \ref{tab:subject-level} presents the subject-level performance of the models, demonstrating improvement in psychological traits estimation when aggregating multiple walking samples. Our model reaches the highest overall weighted F1 Score of 44.6 among the architectures, with GaitFormer obtaining the second highest score, 41.03. The proposed model has an absolute improvement of +7.13 from run-level to subject-level, further proving that certain psychological traits can be better detected when analysing multiple gait samples.

\begin{table}[t]
\caption{MoME performance on auxiliary tasks. Gender and BMI predictions are robust across conditions, while Identification accuracy varies considerably.}

\label{tab:combined_results}
\centering
\resizebox{0.85\linewidth}{!}{
\begin{tabular}{l|l|cccccc|c}
\toprule
\textbf{Scenario} & \textbf{Task} & \textbf{0°} & \textbf{45°} & \textbf{90°} & \textbf{180°} & \textbf{225°} & \textbf{270°} & \textbf{Mean} \\
\midrule

\multirow{3}{*}{NM}
 & Identification & 56.72 & 70.82 & 66.23 & 58.36 & 71.48 & 68.85 & 65.41 \\
 & Gender         & 99.18 & 100.00 & 96.73 & 95.07 & 96.72 & 96.73 & 97.40 \\
 & BMI            & 2.685 & 2.688  & 2.675 & 2.746 & 2.726 & 2.672 & 2.699 \\
\midrule

\multirow{3}{*}{BG}
 & Identification & 42.00 & 54.43 & 58.67 & 56.72 & 64.59 & 64.59 & 56.83 \\
 & Gender         & 98.34 & 95.09 & 98.34 & 93.45 & 95.09 & 95.09 & 95.90 \\
 & BMI            & 2.688 & 2.691 & 2.677 & 2.739 & 2.715 & 2.666 & 2.696 \\
\midrule

\multirow{3}{*}{CL}
 & Identification & 35.41 & 47.54 & 40.00 & 45.25 & 48.52 & 41.31 & 43.01 \\
 & Gender         & 95.09 & 98.36 & 96.72 & 96.72 & 96.72 & 98.36 & 96.99 \\
 & BMI            & 2.687 & 2.693 & 2.675 & 2.739 & 2.714 & 2.667 & 2.696 \\
\midrule

\multirow{3}{*}{WSS}
 & Identification & 45.25 & 65.25 & 59.02 & 46.56 & 61.64 & 54.75 & 55.41 \\
 & Gender         & 98.36 & 100.00 & 96.72 & 95.07 & 98.36 & 96.72 & 97.54 \\
 & BMI            & 2.723 & 2.694  & 2.668 & 2.710 & 2.688 & 2.669 & 2.692 \\
\midrule

\multirow{3}{*}{WSF}
 & Identification & 44.59 & 59.67 & 54.00 & 47.21 & 51.80 & 59.67 & 52.83 \\
 & Gender         & 96.73 & 96.73 & 96.67 & 96.71 & 95.07 & 96.73 & 96.44 \\
 & BMI            & 2.662 & 2.709 & 2.721 & 2.782 & 2.771 & 2.701 & 2.724 \\
\midrule

\multirow{3}{*}{PH}
 & Identification & 45.90 & 51.15 & 45.90 & 48.20 & 56.07 & 51.48 & 49.78 \\
 & Gender         & 95.07 & 98.36 & 95.09 & 95.05 & 93.42 & 96.72 & 95.62 \\
 & BMI            & 2.686 & 2.698 & 2.679 & 2.742 & 2.726 & 2.672 & 2.701 \\
\midrule

\multirow{3}{*}{TXT}
 & Identification & 38.36 & 51.48 & 43.61 & 38.69 & 44.26 & 46.23 & 43.77 \\
 & Gender         & 93.45 & 93.45 & 98.36 & 91.81 & 95.09 & 96.73 & 94.82 \\
 & BMI            & 2.678 & 2.693 & 2.673 & 2.740 & 2.730 & 2.665 & 2.697 \\
\bottomrule
\end{tabular}
}
\end{table}

In this subject-level setting, our model obtains strong results on GHQ (63.33), RSE (59.95), DASS - Stress (57.59), Fatigue (Chronic - 47.67, Acute - 47.39), and DASS - Depression (47.45) showing that these traits have strong correlation with the walking pattern. BFI traits remain the most difficult to predict from movement data, but all of them improve when aggregating multiple viewpoints and scenarios (e.g., Agreeableness goes to 40.33 from 30.76). The improvement in results from run-level to subject-level shows that personality-linked traits emerge more reliably when reducing session noise. 

\subsection{Auxiliary Tasks across Scenarios and Viewpoints}
Our proposed model is also trained to predict other auxiliary tasks in conjunction with the main task of psychological traits estimation. We analyze how its performance on these auxiliary tasks is affected by viewpoints and scenarios. The auxiliary tasks consist of identity recognition, gender prediction, and BMI estimation

For recognition we follow the same evaluation protocol from CASIA-B \cite{yu2006framework}, where the first normal (NM) samples are used as gallery and where the rest are used for probe. Evaluation is performed by comparing each viewing angle against all others, excluding cases with identical viewpoints. The reported accuracy for a specific angle is the result of the average of all scenarios where that angle is utilized in the probe set. We treat gender prediction as a binary classification task and evaluate it with F1 score, and BMI estimation as a regression task and evaluate it with Mean Absolute Error (MAE).

Table \ref{tab:combined_results} shows the performance of the proposed multi-task model across the three additional gait analysis tasks: identification, gender classification, and BMI estimation. The model demonstrates robust gender prediction, consistently achieving 95\% or more in F1 score and with some specific scenarios and angles where it obtains a perfect score. BMI estimation remains relatively similar for all scenarios and angles, with mean absolute errors below 2.8, indicating that gait-based BMI estimation is robust and is unaffected by changes in walking style or viewpoint.

While other works presented identification results on PsyMo only in aggregated form \cite{cosma2025scaling}, we are the first to present extended identification results on PsyMo. We note that identification accuracy varies considerably across scenarios, with the highest performance (65.41\%) being obtained in the normal walking (NM) case. The scenario in which the person is wearing a coat (CL) appears to be the most challenging (43.01\%), which is in line with results on CASIA-B \cite{yu2006framework} on the same walking variation. This probably happens because the coat obstructs a significant part of the body, making it harder to obtain accurate pose estimation results. We observe that walking while texting (TXT) also severely impacts the performance (43.77\%). We believe that this could have two reasons: first, the individual does not move their hands at all in this scenario; second, this is the only scenario where the person does not look in the direction they're walking, affecting their natural gait pattern. Furthermore the TXT and PH variations are so-called "dual-tasks" in which participants perform another action while walking, potentially changing their conscious focus and therefore their walking patterns \cite{dual-tasks}.

\subsection{Ablation on Multi-Task Synergy}
\begin{table}[hbt!]
\caption{Impact of additional activated tasks on psychological traits estimation. Training for multiple tasks simultaneously forces the extraction of meaningful movement features and improves the performance on nearly all tasks.}
\label{tab:additional_tasks}
\centering
\resizebox{\linewidth}{!}{
\begin{tabular}{l|cccc|ccccc}
\toprule
 & \multicolumn{4}{c|}{Activated Tasks} & Identity & Gender & BMI & \multicolumn{2}{c}{Personality} \\
\midrule
Row & \textbf{Id.} & \textbf{Gender} & \textbf{BMI} & \textbf{Personality} & \textbf{Accuracy} & \textbf{F1} & \textbf{MAE} & \textbf{Run F1} & \textbf{Subject F1} \\

\midrule[2pt] 
1 & \checkmark & & & & 50.3 & - & - & - & - \\
\midrule
2 & & \checkmark & & & - & \textbf{96.38} & - & - & - \\
\midrule
3 & & & \checkmark & & - & - & 2.681 & - & - \\
\midrule
4 & & & & \checkmark & - & - & - & 35.25 & 43.29 \\
\midrule[2pt]

5. & \checkmark & & & \checkmark & 48.8 & - & - & 37.47 & 44.23  \\
\midrule
6. & & \checkmark  & & \checkmark  & - & 92.37 & - & 36.41 & 42.95  \\
\midrule
7. & & & \checkmark  & \checkmark  & - & - & 2.718 & 35.93 & 42.61 \\

\midrule[2pt]
8. & \checkmark & \checkmark  & & \checkmark  & 46.92 & 94.62 & - & 37.52 & 43.98  \\
\midrule
9. & \checkmark & & \checkmark  & \checkmark  & 50.2 & - & \textbf{2.65} & 37.45 & 44.28  \\
\midrule
10. & & \checkmark  & \checkmark  & \checkmark  & - & 93.7 & 2.70 & 36.4 & 43.35  \\

\midrule[2pt]
11. & \checkmark & \checkmark  & \checkmark  & \checkmark  & \textbf{52.43} & \textbf{96.38} & 2.70 & \textbf{37.62} & \textbf{44.52}  \\
\bottomrule
\end{tabular}
}
\end{table}
As the proposed model is trained on multiple tasks simultaneously, we are interested in determining how these tasks interact. Consequently, we analyze which tasks enhance performance on the main task of psychological traits estimation and which ones affect it negatively. We train the model with each combination of activated tasks to determine optimal performance and the task interactions.

Table \ref{tab:additional_tasks} shows the performance of training only on 1 task, on 2 tasks at the same time or doing multi-task learning with our proposed model. The results demonstrate the impact of multi-task learning as activating all tasks simultaneously (Row 11) yields the highest performance across nearly all evaluation metrics, achieving 52.43\% identification accuracy, a Personality Run F1 of 37.62\% and a Personality Subject F1 of 44.52\%. This indicates that our model is able to learn to extract various relevant movement information that is complementary in all training objectives.

In particular, we notice that personality prediction consistently benefits from additional activated tasks, mostly from identification. We observe that training personality prediction in conjunction with another task always yields improved results compared to training it individually (Rows 5-7 vs Row 4). Furthermore, performance further increases when training on personality together with two additional tasks (Row 5 vs Row 8; Row 6 vs Row 9; Row 7 vs Row 10). When training on all tasks at the same time, the highest F1 score is obtained for psychological traits estimation. Interestingly, in this setting, identification and gender prediction also reach the highest performance.

\subsection{Expert Specialization}
\begin{figure}[hbt!]
\centering
\includesvg[width=\linewidth]{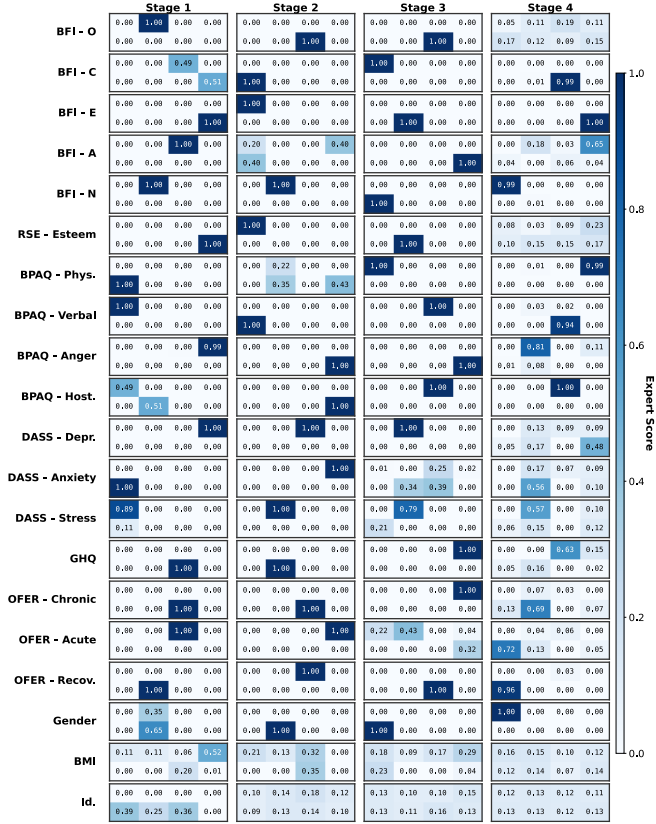}
\caption{Expert activation heatmap across stages and tasks in our proposed MoME architecture. Each cell represents the post-softmax gating weight assigned to a specific expert for a given psychological attribute at a specific stage of movement complexity.}
\label{fig:experts_activation}
\end{figure}

To analyze how tasks interact and how each expert specializes, we plot the expert activations across stages and tasks in Figure \ref{fig:experts_activation}. The heatmap shows the post-softmax gate scores for each task and stage averaged over all test samples. We observe that some tasks rely on a single specialized expert in all stages (e.g., BFI - Neuroticism, BPAQ - Verbal), while others utilize a combination of multiple experts at each stage (e.g., Identification). However, most tasks exhibit a hybrid pattern, utilizing specialized experts in some stages and multiple experts in others.

Certain tasks are similar in terms of the experts utilized across the stages, suggesting that the model identified similarities in the gait patterns associated with these psychological markers. For example, BFI - Extraversion and RSE - Esteem share the same specialized expert across the first 3 stages, while having only a slight overlap in Stage 4 in Expert 8. This is consisted with prior psychological research reporting a moderate correlation between extraversion and self-esteem \cite{robins2001personality}, and with some works suggesting that both may be linked to positive emotionality \cite{deneve1998}. 

Another example is the GHQ - OFER Chronic pair of tasks which employ the same expert in Stages 1 and 3, with a slight overlap in Stage 4 and entirely different specialized experts in Stage 2. This is in line with the work of Lawrie et al. \cite{ghq-fatigue}, which reported a strong correlation between Fatigue levels and scores on the General Health Questionnaire.

We also observe task pairs that have a degree of expert overlap, though their psychological association remains uncertain or weakly supported. 
One example is our model's association of the BFI Agreeableness trait with the OFER Acute, utilizing the same expert in Stage 1 and showing a varying degree of overlap in the subsequent stages. This association remains uncertain, as some studies have reported a link between high Agreeableness and lower fatigue, while others have found the opposite pattern, and several have observed no significant relationship \cite{stephan2022personality}.

An interesting phenomenon appears in the identification task which allocates weight across many experts rather than collapsing onto one. We hypothesize that this approach is a result of the training objective. Unlike the other tasks which predict a specific class label or scalar, identification is the only task trained with triplet loss, optimizing the embedding space directly. The maximization of this objective is best achieved by drawing on complementary components, which are extracted by different experts at each stage and provide more independent directions to tell identities apart.

\section{Limitations and Societal Impact}
In our results, certain attributes (e.g., BFI and BPAQ subscales) remain particularly challenging to predict, which indicates that inferring fine-grained psychological attributes from gait alone is difficult and predictions should be interpreted with caution. Subject-level aggregation yields a +7.13 absolute gain over run-level, implying that single-run predictions are inherently noisy. While our averaging strategy improves results, this design choice limits applicability in settings where only one sequence is available.

Another important limitation concerns the nature of the data utilized. All the compared methods are trained and evaluated in data collected in controlled laboratory settings, consisting primarily of walking videos of students of similar age from Romania. As a result, the method may not generalize across age groups, cultural backgrounds or environmental conditions. Furthermore, factors that are critical in real-world deployment, such as camera height, lighting and occlusions, may also limit performance.

Finally, societal and ethical implications must be acknowledged. Inferring psychological attributes from gait can expose sensitive information that individuals do not intend to reveal. This raises significant concerns regarding privacy, consent, and possibility of misuse. We acknowledge that such a technology could be utilized for discriminatory or manipulative reasons. For these reasons, the proposed method is designed for research, not intended for practical applications. 

\section{Conclusions}
This work investigates the task of psychological traits estimation from gait in the context of multi-task learning and introduces a Multi-Stage \textbf{M}ixture of \textbf{M}ovement \textbf{E}xperts (MoME) architecture for solving the problem. By processing the walking pattern through progressively complex stages of movement and employing task-specific gating across expert modules, our approach demonstrates strong performance in multi-task learning from motion data.

In particular, our proposed model achieves state-of-the-art performance, outperforming widely utilized gait analysis architectures that employ skeleton, silhouette or mesh data on the PsyMo benchmark across 17 different psychological traits. On the run-level evaluation, MoME obtains 37.47\% weighted F1 score, whereas on the subject-level evaluation it achieves 44.6\%. Our results highlight which psychological traits show strong correlations with movement and which remain challenging to infer from gait alone.

Beyond psychological attributes, we demonstrate that our proposed model is adaptable to auxiliary tasks such as re-identification, gender prediction and BMI estimation from gait data. Our analysis shows that these additional tasks improve overall psychological traits estimation by encouraging the extraction of meaningful movement features. Overall, our work provides a strong foundation for multi-task gait analysis, enabling advances in psychological assessment and behavioural biometrics.

{\small
\bibliographystyle{ieee}
\bibliography{bibliography}
}

\end{document}